\documentclass[sn-mathphys]{sn-jnl}

\usepackage{amssymb}

\usepackage{float}
\usepackage{amsthm}
\usepackage{booktabs}
\usepackage{caption}
\usepackage{subfigure}
\usepackage[shortlabels]{enumitem}
\usepackage[final]{pdfpages}
\usepackage{bm}
\usepackage{algorithmicx,algorithm}
\usepackage{threeparttable,color}
\usepackage[title]{appendix}
\usepackage{mathtools,xparse}
\usepackage{amsmath,upgreek,bm}
\usepackage{multirow}
\usepackage[english]{babel}
\usepackage{graphicx}

\jyear{2021}%

\theoremstyle{thmstyleone}%
\theoremstyle{thmstyletwo}%

\theoremstyle{thmstylethree}%

\begin{document}

\title{A Robust graph attention network with dynamic adjusted Graph}





\author[1]{\fnm{Xianchen} \sur{Zhou}}
\email{zhouxianchen13@nudt.edu.cn}

\author[2]{\fnm{Yaoyun} \sur{Zeng}}
\email{yaoyun\_zeng@nudt.edu.cn}
\author[3]{\fnm{Zepeng} \sur{Hao}}
\email{haozepeng@gmail.com}

\author*[4]{\fnm{Hongxia} \sur{Wang}}
\email{wanghongxia@nudt.edu.cn}

\affil[1]{\orgname{National University of Defense Technology}, \orgaddress{\city{Changsha}, \postcode{410072}, \state{Hunan}, \country{China}}}


\abstract{Graph Attention Networks(GATs) are useful deep learning models to deal with the graph data. However, recent works show that the classical GAT is vulnerable to adversarial attacks. It degrades dramatically with slight perturbations. Therefore, how to enhance the robustness of GAT is a critical problem.

Robust GAT(RoGAT) is proposed in this paper to improve the robustness of GAT  based on the revision of the attention mechanism. Different from the original GAT, which uses the attention mechanism for different edges but is still sensitive to the perturbation,  RoGAT adds an extra dynamic attention score progressively and improves the robustness. Firstly, RoGAT revises the edge‘s weight based on the smoothness assumption which is quite common for ordinary graphs. Secondly, RoGAT further revises the features to suppress features' noise. Then, an extra attention score is generated by the dynamic edge's weight and can be used to  reduce the impact of adversarial attacks. Different experiments against targeted and untargeted attacks on citation data on citation data demonstrate that RoGAT outperforms most of the recent defensive methods.}


\maketitle
\section{Introduction}
Non-Euclid data occurs widely in our daily life and Graph Attention Network(GAT)~\cite{ve2018graph} achieves remarkable performance in these data represented by graphs. It generates node embedding by using a local aggregation function~\cite{Scarselli2009The,gilmer2017neural,NIPS2017_6703} with attention mechanism~\cite{vaswani2017attention,bahdanau2014neural}, which computes the hidden representation by the features of connected nodes with different attention weight coefficients. The attention mechanism makes
GAT focus on the relevant part and perform well. However, GAT is vulnerable to the adversarial attacks, which means subtle perturbations may degrade its performance significantly. The lack of robustness makes GAT practical limited in several fields like military and  finance with high requirement of security. For instance, in the social secure field, the criminal can create or hide some social relationship to escape the examination of GAT. 
Hence, developing the robustness of GAT to resist the different kinds of adversarial attacks~\cite{jin2020adversarial} is important and urgent.

\begin{figure}
   \centering
   \includegraphics[width=1\linewidth]{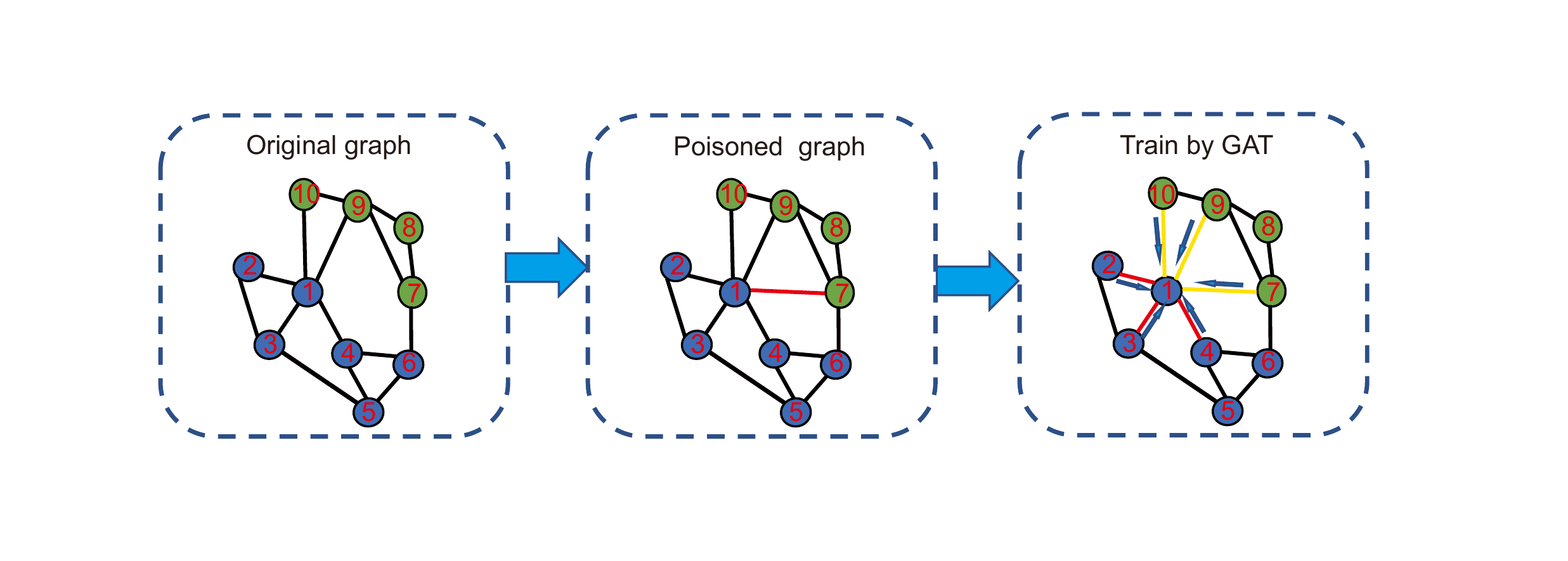}
   \caption{Original graphs can be attacked by the negative edges between node 1 and 7 with different labels. The update function of node 1 aggregates more negative information from nodes 7, 9, 10 than positive information from nodes 2, 3, 4.}
   \label{fig:twoexample}
 \end{figure}

The vulnerability of GAT is due to its aggregation function. 
 As is shown in Figure \ref{fig:twoexample},
 the aggregation function aggregates the information from different kinds of neighbors. In intuition, information from the  neighbors of similar or same labels(positive edges) make positive effects, while information from  dissimilar or distinct neighbors(negative edges) may make negative effects on the iteration of node features. However, adversarial attacks add extra negative edges or delete positive edges, which degrades GAT.
In this paper, we assume  graphs to be analyzed satisfies feature smoothness assumption~\cite{wang2019knowledge} which is common in most graphs. It means for most nodes in graph, there often exists more neighbors with positive edges  than that with negative edges. Note that there also exists some  heterogeneous graphs which don't satisfy smoothness assumption, we do not discuss these graphs here and will address them in the future. We aim to design an improved GAT to defend the adversarial attacks based on the prior information of graphs.
  
An intuitive idea of defensive technology is using prior information to increase the positive effects and reduce negative effects of neighbors.  Two problems are faced here: (1) What kind of methods  helps us to distinguish two kinds of neighbors? (2) How to design the attention scores for different edges? 
This paper proposes a Robust GAT(RoGAT) to solve these problems. RoGAT distinguishes the positive and negative  neighbors based on the Laplacian regularization~\cite{wang2019knowledge} and designs an extra dynamic scores to adjust the attention effects for different edges.

The contribution of RoGAT can be summarized as follows.

(1) The mode defend against adversarial attacks by increasing extra edge attention scores to distinguish two kinds from adversarial edges based on the feature smoothness assumption~\cite{wang2019knowledge}.

(2) The model adjusts the graph structure and feature iteratively during the training procedure, which leads to the ratio of attention scores between real edges and adversarial edges increases. 

(3) The experiments on various  real-world graphs show that RoGAT can adjust the ratio of attention scores between negative and positive edges iteratively and thus outperform other defensive methods in the node classification task under different types of attacks. 

The implementation of RoGAT is based on the DeepRobust~\cite{li2020deeprobust} repository for adversarial attacks and the experimental settings to reproduce our results can be found in  https://github.com/zhouxianchen/robustGAT. The rest of the paper is organized as follows.
 Section 2 gives the notations and discusses related works of adversarial attacks and defensive methods. Section 3 reviews the original GAT and . In section 4, the relative merits of GAT is discussed and RoGAT is proposed to improve the performance of defending attacks. Section 5 gives some experiments to verify the conclusion.  Section 6 gives a further discussion and conclusion of our methods. The last section is the acknowledgement.

\section{Related works and Notations}

\subsection{Related Works}
Recently, there is some research about adversarial attacks and defense on Graph Neural Networks, which can be used in GAT.
The graph adversarial attacks can be divided into targeted attacks and untargeted attacks. The targeted attacks like nettack~\cite{zugner2018adversarial} and RL-S2V~\cite{dai2018adversarial} tend to let the trained
model misclassify a small set of test samples, while untargeted attacks like metattack~\cite{zugner_adversarial_2019} aimed to let the trained model have bad overall performance on all test data. Nettack introduces the unnoticeable perturbations on both structures and features. RL-S2V uses reinforcement learning to generate attacks on GNNs. The metattack parameterized the graph structure and used the gradient information to attack GAT. 

As to the method of defending the adversarial attacks, 
one perspective to achieve robustness is to eliminate the influence of perturbations such as adding or removing the adversarial edges or clearing up the change of node features. The criteria of eliminating the influence are mainly based on the  prior information of the graph in specific applications. 
For example, Wu~\cite{wu2019adversarial} applied the Jaccard similarity to eliminate the edges between nodes with low similarity. 
GNNguard~\cite{zhang2020gnnguard} can detect and quantify the relationship between the graph structure and node features based on  the hypothesis that similar nodes are more likely to interact than dissimilar nodes. It estimated an importance weight for every edge to reduce the influence of fake edges.
RGCN~\cite{10.1145/3292500.3330851} added the penalization of adversarial edges and modeled the hidden layers by Gaussian distributions to reduce the effect of attacks. 
PTDNet~\cite{luo2021learning} uses nuclear norm regularization  to drop some task-irrelevant edges and improve the robustness.
ProGNN~\cite{jin2020graph} assumed that the graph should be low-rank and sparse and then gave a progressive model for adversarial training. These models used the poisoned graph for training and estimated the clean graph by prior information~\cite{tang2020transferring,10.1145/3292500.3330851}.
Besides,  another perspective tries to figure out the pattern  of adversarial attacks and design defensive methods. 
Wei~\cite{jin2021node} found that adversarial attacks can destroy graph structure.They proposed SimPGCN which can effectively and efficiently preserve node similarity while exploiting graph structure. In GCNSVD~\cite{10.1145/3336191.3371789}, it can defend the metattack~\cite{zugner2019adversarial} by reducing the rank of the adjacency matrix.
PA-GNN~\cite{tang2020transferring} designed a meta-optimization algorithm by penalizing the perturbations to restrict the negative impact of adversarial edges.  HSC-GAT~\cite{zhao2021robust} proposes a holistic semantic constraint GAT which approaches the joint modeling of graphs to mitigate the perturbations. These methods can defend the adversarial attacks under the different situations.

\subsection{Notations}
Let $G=(V,E)$ be a graph, where $V=\{v_1,v_2,\cdots,v_N\}$ is the set of nodes and $E$ is  the set of edges. Each graph can be represented by the adjacency matrix $A\in \mathbb{R}^{N\times N}$, where  $a_{ij}$, the  $(i,j)$th elements of $A$, represents the link weight of node $v_i$ and $v_j$. In addition, $X = [x_1,x_2,\cdots, x_N]^{\top}\in \mathbb{R}^{N\times d}$ denotes the feature matrix where $x_i\in \mathbb{R}^d$ is the feature vector of $v_i$. Hence a simple representation of a graph is $G=(V,A,X)$.

Here we consider the semi-supervised node classification problem. Only parts of nodes $\mathcal{V}_p=\{v_1,v_2,\cdots,v_m\},m<N$ are annotated. $\mathcal{Y}_p=\{y_1,y_2,\cdots,y_m\}$, where $y_i$ is the label of $v_i$. Given graph $G=(V,A,X)$ with partial label $\mathcal{Y}_p$,  the goal  of node classification is to predict the labels of unlabeled nodes.

\section{GAT}
 GAT generates the new feature of one node by combining the feature vector of each node with attention in its neighbors. The attention mechanisms make GAT more flexible in aggregation. The formulation of $L$-layer GAT can be denoted by

\begin{equation}\label{aggregation}
\begin{aligned}
& x_{v}^{(0)}=x_v, v\in\{1,\cdots,N\},\\
& x_{v}^{(k)}=\sigma\left(\sum_{u \in \mathcal{N}(v)} \alpha_{u v}^{(k-1)} W^{(k-1)} x_{u}^{(k-1)}\right), k = 1,2,\cdots,L
\end{aligned}
\end{equation}
where 
\begin{equation}\label{attention:1}
\alpha_{uv}^{(k-1)}=\frac{\exp \left(\operatorname{Leaky} \operatorname{ReLU}\left(\overrightarrow{\mathbf{a}_k}^{T}\left[W^{(k-1)}  x_{v}^{(k-1)} \| W^{(k-1)}  x_{u}^{(k-1)}\right]\right)\right)}{\sum_{l \in \mathcal{N}{(v)}} \exp \left(\operatorname{LeakyReLU}\left(\overrightarrow{\mathbf{a}_k}^{T}\left[W^{(k-1)}  x_{v}^{(k-1)} \| W^{(k-1)}  x_{l}^{(k-1)}\right]\right)\right)}.
\end{equation}
$W^{(k)}\in \mathbb{R}^{d_k\times d_{k+1}}$ are the parameters to learn and $\sigma$ is the activation function. $d_0=d$, $d_L=K$ is the number of class. $\|$ represents a concatenation operator,  $\overrightarrow{\mathbf{a}_k} \in \mathbb{R}^{2d_{k}}$ is a weight vector multiplying the concatenation vector, $\mathcal{N}(v)$ represents the neighbors of node $v$.

  The multi-head attention is used to improve the performance of GAT:

  \begin{equation}
x_{v}^{(k)}=\|_{p=1}^{M} \sigma\left(\sum_{u \in \mathcal{N}(v)} (\alpha_{u v}^{(k-1)} W^{(k-1)} )^p x_{u}^{(k-1)}\right), k=1,2,\cdots,L
\end{equation}
where  $x_v^{(0)}=x_v$, $v\in\{1,2,\cdots,N\}$. The update feature $x_v^{(k)}$ relies on the neighbors' features $x_u^{(k-1)}$ and the weights $\alpha_{ij}$ are computed according to the features.

Denote the learning parameters of GAT by $\theta=\{W^{(0)},\cdots,W^{(L-1)}\}$  which including all the $W^{(k)}$ in each layer. Then for node classification problem, GAT learns a function $f_\theta^{\text{GAT}}:\mathcal{V} \rightarrow \mathcal{Y}$ by applying Boolean classification function to $x_v^{(L)}$ to predict unlabeled nodes. And the objective function is the sum of loss for the labeled nodes,
\begin{equation}
\label{loss}
\begin{aligned}
\mathcal{L}_\text{GAT}(\theta,A,X,\mathcal{Y}_p)= \sum_{v_{i} \in \mathcal{V}_{p}} \ell (f_{\theta}(X,  A)_{i}, y_{i})\\
  \theta^* = \arg \min _{\theta}\mathcal{L}_\text{GAT}(\theta,A,X,\mathcal{Y}_p),
\end{aligned}
\end{equation}
where $\theta$ is the parameters of GAT, $f_{\theta^*}^{\text{GAT}}(X,  A)_i$ is the predicted label of $v_i$.

\section{The proposed method}

\subsection{Analysis of GAT}
%

    To figure out the relationship between graph structure and accuracy of GAT,we established an experiment both on the simulated and real graphs. 

As to the simulated data, one thousand nodes are labeled by two classes equally. Any two nodes are connected with probability $p_1$ if they are in the same class and probability $p_2$ while in the distinct class. 
The number $N_1$ and $N_2$ represent the links between the same and distinct labels respectively. 
10\% percent of nodes are set as the training nodes, while another 10\% of nodes are chosen randomly as the testing nodes.
 We compare the accuracy of GAT  on random graphs with different numbers $N_1/N_2$.

As is shown in Table \ref{ratio_table}, when $N_1$ approximates $N_2$,  the accuracy of GAT degrades to 50\%. GAT performs better when the ratio increases. And for Cora, Citeseer and Polblogs, the performance of GAT is also related to $N_1$/$N_2$.  

 In fact, the aggregation function for GAT is tightly connected with graph structure and features~\cite{dai2018adversarial,wu2019adversarial}.  Although GAT aggregates the hidden features of its neighbors with attention scores, it cannot distinguish the information came from positive or negative neighbors well. When the labels in neighbors tend to the same, the propagation of GAT preserves the feature well. While the labels in neighbors have multiple labels, the features that came from different labels make  considerable and negative effects on the iterated feature. 
 Therefore, $N_1/N_2$ affects the aggregation process of all nodes averagely.

\begin{table}[htbp]
  \centering
  
  \caption{The accuracy of GAT on random graph and three datasets with different link ratio($N_1$ and $N_2$ represents the number of links between the same and different labels, respectively)}.
  \begin{tabular}{@{}ccccc@{}}
    \toprule
    \multicolumn{1}{l}{Dataset}   & \multicolumn{1}{l}{$N_1$} & \multicolumn{1}{l}{$N_2$} & \multicolumn{1}{l}{$\frac{N_1}{N_2}$} & \multicolumn{1}{l}{GAT peformance} \\ \midrule
    \multirow{4}{*}{Random graph} & 4982                      & 4864                      & 1                                  & 0.5150                             \\
    & 9894                      & 4818                      & 2                                  & 0.8125                             \\
    & 14774                     & 5076                      & 3                                  & 0.9637                             \\
    & 19714                     & 4944                      & 4                                  & 0.9988                             \\  \midrule
    Cora                          & 8152                      & 1986                      & 4.10                               & 0.8397                             \\ 
    \midrule
    Citeseer                      & 5402                      & 1934                      & 2.79                               & 0.7326                             \\
    \midrule
    Polblogs                      & 30278                     & 3150                      & 9.61                               & 0.9535                             \\ \bottomrule
  \end{tabular}
  \label{ratio_table}
\end{table}

 Therefore, some attack methods degrade GAT by affecting $N_1/N_2$ of graphs based on various technologies. Since GAT uses the attention mechanisms to mix various features, the intuition is that GAT can adjust the attention scores for useful and useless neighbors' information automatically to resist the attacks during the training procedure. However,  Table \ref{ratio_table22} displays the $\frac{N_1}{N_2}$of Cora and accuracy of GAT under different perturbations rates.
 The adversarial attack can add or delete the edges to change $N_1/N_2$ and degrade GAT significantly~\cite{10.1145/3292500.3330851,jin2020graph}.

Nevertheless, for a given graph with most unlabeled nodes, it is impossible to change $N_1/N_2$ easily. In this paper, we tends to enhance the impact of $N_1$ positive edges and reduce that of $N_2$ negative edges.

\begin{table}[htbp]
  \centering
  
  \caption{The accuracy of GAT on Cora dataset with different perturbations rates with mettack.}
  \begin{tabular}{llllll}
    \cline{1-5} \cline{5-5}
    Perturbation rates      & $N_1$ &  $N_2$ & $\frac{N_1}{N_2}$ & GAT accuracy  \\ \cline{1-5} \cline{5-5} 
    0 & 8152                          & 1986                          & 4.10              & 0.8397     \\
    0.05 & 8232                                  & 2412                         & 3.41              & 0.8044         \\
    0.10& 8294                               & 2840                            & 2.92             &   0.7561     \\
    0.15 & 8286                               & 3332                           & 2.49              & 0.6978         \\
    0.20         & 8228                              & 3852                         & 2.13           & 0.5994   \\
    0.25     & 8252                                & 4240                           & 1.95           & 0.5478        \\  \bottomrule
  \end{tabular}
  \label{ratio_table22}
\end{table}

\subsection{RoGAT}
\label{sec RoGAT}
 Note that the original attention scores in GAT are computed based on the node feature and labels of neighbors~\cite{zhang2019adaptive}. And the graph structure only decides the choice of neighbors but cannot help to adjust the attention scores. Adversarial attacks add negative edges or reduce the positive edges to affect the aggregation neighbors.Therefore, we tend to revise the aggregation function and insert an extra attention score to reduce the effect of adversarial edges.
 Intuitively, it should assign relatively small attention scores to those negative edges though the prior information. Note that the proper attention score is related not only to structure $A$ but also to feature $X$, we establish a robust model which adjusts the graph structure and feature to generate an extra score, which helps to reduce the negative edges bringing by adversarial attacks.

\subsubsection{The optimization model}
In most situations especially for homogeneous networks, the connected nodes with the same labels in a graph tend to share similar features. For example, in citation networks, the entities with similar bag-of-words features tend to connect and belong to the same class~\cite{kipf2017semi}. And two connected individuals in social graphs may share similar features since they tend to have related hobbies or characters~\cite{mcpherson2001birds}. 
The characteristic can be represented by the feature smoothness regularization $\operatorname{tr}\left(X^{\top}LX\right)$, where $L$ is the weighted Laplacian matrix of the graph. Adjust the effect to enlarge the 
We propose the following optimization model that obtains the revised structure $\bar{A}$, feature $\bar{X}$ and parameters $\theta$, which can be described as:

\begin{equation}
\label{optim}
\begin{aligned}
(\bar{A}^*,\bar{X}^*,\theta^*) &= \arg \min _{\theta,\bar{A}\in \bar{\mathcal{A}},\bar{X}\in \mathcal{X}} \mathcal{L}_{\text{re}}(\bar{A},\bar{X},\theta)+\lambda \mathcal{L}_\text{RoGAT}(\theta,\bar{A},\bar{X},\mathcal{Y}_L)\\  \text{with} &\quad \quad \quad \\
\mathcal{L}_{\text{re}}(\bar{A},\bar{X},\theta)&=	\|A-\bar{A}\|^2+\beta\|X-\bar{X}\|^2+\alpha \operatorname{tr}\left(\bar{X}^{\top}\bar{L}\bar{X}\right)\\&=
\|A-\bar{A}\|^2+\beta\|X-\bar{X}\|^2+\frac{\alpha}{2} \sum_{i, j=1}^{N} \bar{A}_{i j}\left(\bar{\mathbf{x}}_{i}-\bar{\mathbf{x}}_{j}\right)^{2},
\end{aligned}
\end{equation}
where $\mathcal{L}_\text{RoGAT}(\theta,\bar{A},\bar{X},\mathcal{Y}_L)$ is given by
\begin{equation}
    \mathcal{L}_\text{RoGAT}(\theta,A,X,\mathcal{Y}_p)= \sum_{v_{i} \in \mathcal{V}_{p}} \ell (f_{\theta}^{\text{RoGAT}}(X,  A)_{i}, y_{i})
\end{equation}
\eqref{loss}, $\alpha,\beta$,$\gamma$ are non-negative parameters. $\bar{L}=\bar{D}-\bar{A}$ is the laplacian matrix whose diagonal element $\bar{D}_{ii}=\Sigma_j {\bar{A}_{ij}}$. $\bar{\mathcal{A}}$ represents the domain of adjacency matrix, with element ranging from 0 to 1, while $\bar{\mathcal{X}}$ represents the domain of feature matrix. 

Different from the original GAT, $f_\theta^{\text{RoGAT}}:\mathcal{V} \rightarrow \mathcal{Y}$  revises the \eqref{aggregation} by inserting modified attention score determined  by the revised structure $\bar{A}$. The  revised aggregation function can be written as	
\begin{equation}
x_{v}^{(k)}=\|_{m=1}^{M} \sigma\left(\sum_{u \in \mathcal{N}(v)} (\bar{\alpha}_{u v}^{(k-1)} W^{(k-1)})^m x_{u}^{(k-1)}\right),
\end{equation}
where $\bar{\alpha} _{uv}$ is a modified attention combining the feature attention \eqref{attention:1} and graph structure attention obtained by the optimization model \eqref{optim}. The revised attention is defined as
$$\bar{ \alpha} _{uv} = \bar{A}_{uv} \alpha_{uv}.$$Here $\bar{A}_{uv}$ is the link weight of the current $\bar{A}$ computed by optimization model.
\subsubsection{The optimization algorithm}
We update the graph structure $\bar{A}$ and feature $\bar{X}$ and the parameters of GAT alternatively to solve the optimization model \eqref{optim} as follows.
To solve problem \eqref{optim}, firstly we fix $\bar{X}$ and consider the update of $\bar{A}$ by
  \begin{equation}\label{target}
\begin{aligned}
\underset{\bar{A} \in \bar{\mathcal{A}}}\min \|A-\bar{A}\|^2+ \alpha \operatorname{tr}\left(\bar{X}^{\top}\bar{L}\bar{X}\right).
\end{aligned}
\end{equation}
We initialize $\bar{A} = A$, then 
update $\bar{A}$ by using projected gradient descent method:

\begin{equation}
\bar{A}\longleftarrow  P_\mathcal{A}(\bar{A}-\eta_1 \mathcal{L}_s)
=
\bar{A}-\eta_1 \nabla_{\bar{A}}( \|A-\bar{A}\|^2+ \alpha \operatorname{tr}\left(\bar{X}^{\top}\bar{L}\bar{X}\right)),
\end{equation}
where
\begin{equation}
P_\mathcal{A}(\bar{A})= \left\{
\begin{aligned}
0, \bar{A_{ij}}<0\\
1, \bar{A_{ij}}>1\\
A, \text{otherwise.}
\end{aligned}
\right.
\end{equation}

Then we fix $\bar{A}$ and consider the update of $\bar{X}$ by
\begin{equation}\label{target2}
\begin{aligned}
\underset{\bar{X} \in \bar{\mathcal{X}}}{\min}\|X-\bar{X}\|^2+ \gamma \operatorname{tr}\left(\bar{X}^{\top}\bar{L}\bar{X}\right)+ \lambda \mathcal{L}_\text{RoGAT}(\theta,\bar{A},\bar{X},\mathcal{Y}_p).
\end{aligned}
\end{equation}
We use gradient descent method to update $\bar{X}$:
\begin{equation}
\bar{X}\longleftarrow  P_\mathcal{X} {(\bar{X}-\eta_2  \nabla_{\bar{X}} ( \|X-\bar{X}\|^2+ \gamma \operatorname{tr}\left(\bar{X}^{\top}\bar{L}\bar{X}\right)+ \lambda \mathcal{L}_\text{GAT}(\theta,\bar{A},\bar{X},\mathcal{Y}_p)))}.
\end{equation}
where $P_\mathcal{X}(\cdot)$ is the projection on the feature matrix domain.

%

The solving procedure of model \eqref{optim} can be divided into the update of graph structure, feature and parameters of RoGAT alternatively. Algorithm \ref{algorithm1} gives the iteration procedure.


    


  
\begin{algorithm}[ht]
  \caption{RoGAT} 
  {\bf Input:} 
  Graph $G=(V,A,X)$ and part of nodes $V_p$ with labels $\mathcal{Y}_p $.
\\
{\bf Parameters:}  \\$\alpha$, $\gamma$, $\lambda$: the non-negative parameters\\
  $T_1$,$T_2$: outer and inner maximum iteration steps\\
   $\eta_1, \eta_2,\eta$: learning rates for sub-optimization problems.
\\
 {\bf Output:} 
the RoGAT model with learned parameters $\theta$.
  \begin{algorithmic}[1]	
    \State Initialize the RoGAT model with given structure $A$ and set $\bar{A} \longleftarrow A$, $\bar{X} \longleftarrow X$.
    
    \State Randomly initialize the parameter $\theta$ of RoGAT.

\For {$i=1$ to $T_1$}:
          \State	$\bar{A}\longleftarrow \bar{A}-\eta_1 \nabla_{\bar{A}}( \|A-\bar{A}\|^2+ \alpha \operatorname{tr}\left(\bar{X}^{\top}\bar{L}\bar{X}\right)),$
          \State	$\bar{A}\longleftarrow P_\mathcal{A}(\bar{A}),$
          \State $\bar{X}\longleftarrow \bar{X}-\eta_2 \nabla_{\bar{X}}(\|X-\bar{X}\|^2+ \gamma \operatorname{tr}\left(\bar{X}^{\top}\bar{L}\bar{X}\right)+ \lambda \mathcal{L}_\text{RoGAT}(\theta,\bar{A},\bar{X},\mathcal{Y}_p)),$
           \State	$\bar{X}\longleftarrow P_\mathcal{X}(\bar{X}),$
          \For{$i=1$ to $T_2$}:
    
    \State $\theta \leftarrow \theta  - \eta\frac{\partial \mathcal{L}_{\text{RoGAT}}\left(\theta, \bar{A}, \bar{X}, y_{p}\right)}{\partial \theta}.$

          \EndFor
  \EndFor
  
  \State Return $\theta$ and RoGAT.
  \end{algorithmic}
\label{algorithm1}
\end{algorithm}


%
%
\section{Experiments}  

%

In this section, we firstly empirically evaluate RoGAT on semi-supervised problems
 with the state of the art defense methods under different kinds of adversarial attacks.
Then we analyze the effect of parameters and explain why our method works. 
\subsection{Experimental settings}
\subsubsection{Experimental datasets}
We choose three benchmark datasets Cora, Citeseer and Polblogs as ~\cite{zugner2018adversarial,zugner2019adversarial}.
The largest component of these datasets ~\cite{10.1145/3336191.3371789,jin2020graph} are used in this paper in Table \ref{dataset} .

\begin{table}[htbp]
\caption{Datasets}
\label{dataset}
\begin{equation*}
\begin{array}{c|cccc}\hline & \mathrm{N}_{\mathrm{LCC}} & \mathrm{E}_{\mathrm{LCC}} & \text { Classes } & \text { Features } \\ \hline \text { Cora } & 2,485 & 5,069 & 7 & 1,433 \\ \text { Citeseer } & 2,110 & 3,668 & 6 & 3,703 

\\   \text { Polblogs } & 1,222 & 16,714 & 2 & / \\ \hline\end{array}
\end{equation*}
\end{table}
\subsubsection{Baselines}

Here we compare RoGAT with different Graph Neural Networks and implement the defensive models by the DeepRobust library~\cite{li2020deeprobust}.
\begin{itemize}
\item \textbf{GCN}~\cite{kipf2017semi}:  The classical and widely used GCN defines the graph convolution in a spectral domain.
\item \textbf{GAT}~\cite{ve2018graph}: GAT uses the attention mechanism to learn the representation of nodes.
\item \textbf{RGCN}~\cite{10.1145/3292500.3330851}: RGCN assumes that all the node representations are defined by Gauss distributions and uses an attention mechanism to reduce the influence of the nodes with high variance.
\item \textbf{GCN-Jaccard}~\cite{wu2019adversarial} : As attacks tend to link the nodes with huge feature differences, GCN-Jaccard makes a judgment to eliminate part of edges between nodes with small similarities. 
\item \textbf{GCN-SVD}~\cite{10.1145/3336191.3371789}: Since \textit{nettack} is a high-rank  attack, GCN-SVD uses a low-rank approximation of the perturbed graph for further training. This model can also be extended to  non-targeted  and random attacks.
\item \textbf{ProGNN}~\cite{jin2020graph}: ProGNN assumes that graph data in reality is low-rank and sparse. It uses the progressive procedure to adjust the structure and parameters of  GCN. This method performs robustly under three kinds of attacks but is time-consuming.
\item \textbf{ADA-UGNN}~\cite{ma2021unified}: A general GNN frameworks which is suited for handling varying smoothness properties. 
\item \textbf{HSC-GAT}~\cite{zhao2021robust}: A holistic semantic constraint GAT which approaches the joint modeling of graphs to mitigate the perturbations.

\end{itemize}

\subsubsection{Parameter settings}
Since RoGAT is based on GAT, we choose the default settings about GAT in~\cite{ve2018graph} with a two-layer  model. 
Here the dropout parameter $p=0.6$ is applied to both layers' input.  The learning rate for the training feature and adjacency matrix for SGD is set by $0.01$.  For GCN, we use the default settings in~\cite{kipf2017semi}. For RGCN, we use the same settings as the experiments in~\cite{jin2020graph} with $\{16, 32, 64, 128\}$ hidden units. For GCN-Jaccard, $\{0.01,0.02,0.03,0.04,0.05,0.1\}$ are set as the threshold of similarity for removing the edges for different perturbations ratios. For GCN-SVD, $\{5,10,15,50,100,200\} $ are used as the reduced rank. 

For all the tested graphs, we randomly choose 10\% of nodes as the training datasets and 10\% of nodes as the validation datasets. The remaining 80\% of nodes are used for testing for the  non-targeted attack. The inner and outer iterations $T_1$ and $T_2$ are set by 10. The learning rate $\eta_1$ and $\eta_2$ are set by 0.01.The other hyper-parameters are selected by the accuracy of the validation and manual test. All the experiments are executed 10 times with different random seeds. 

\subsection{Defensive performance}
\subsubsection{Under the non-targeted adversarial attack}
First, we evaluate the performance of RoGAT against the non-targeted adversarial attack, which aims to degrade the performance on all nodes. Here we use the metattack as the non-targeted attack and adopt the same parameter settings as~\cite{zugner2019adversarial}. The Meta-self  attack for Cora, CIteseer, and Polblogs is considered  the most effective attack. As is shown in Table \ref{table1}, we compare RoGAT with the other six methods and vary the perturbations rate from 0\% to 25\%. All the experiments are conducted 10 times, and then the average accuracy and standard deviation are recorded.  RoGAT performs the best under the meta attack for all the tested datasets.
\begin{itemize}
\item RoGAT outperforms other methods almost for all the perturbations ratios in Cora and Citeseer datasets and has better performance for larger perturbations in the polblogs dataset.  Specifically, the classification accuracy of RoGAT when processing the Cora and Citeseer datasets with 25\% disturbance is 13\% and 2\% higher than other methods, respectively. In addition, for the Polblogs dataset, under 15\% to 25\% interference, the performance of RoGAT is better than other methods by 2\% to 15\%. 

\item Although ProGNN has good performance when dealing with Cora and Citeseer under the larger ratio of perturbations, RoGAT performs best. Compared with ProGNN, RoGAT ignore the regularization of sparsity and low-rank but can still adapt well to the non-targeted adversarial attack. It means that the attention  mechanism of RoGAT, which considers the feature smoothing can adjust the weight of fake and real edges.
\item Compared with ProGNN, RoGAT has a lower computation time without the regularization of sparsity. ProGNN costs more than 15 minutes for one script in cora with 2080Ti GPU, while the average time for one RoGAT training is 20 seconds.
\end{itemize}

\begin{table}[htbp]

\caption{Node classification performance (Accuracy$\pm$Std) under non-targeted attacks(metattack)}
\resizebox{\textwidth}{40mm}{
  \begin{tabular}{c|c|cccccccc}
    \toprule
    Dataset                   & Ptb (\%) & GCN~\cite{kipf2017semi}                                      & GAT~\cite{ve2018graph}                                    & RGCN~\cite{10.1145/3292500.3330851}           & GCN-Jaccard~\cite{wu2019adversarial}  & GCN-SVD~\cite{10.1145/3336191.3371789}       & Pro-GNN~\cite{jin2020graph}  &ADA-UGNN\cite{ma2021unified}   & \textbf{RoGAT}                            \\ \cline{1-9}
    \multirow{7}{*}{Cora}     & 0             & 83.50$\pm$0.44                           & 84.57$\pm$0.65 & 83.09$\pm$0.44 & 82.05$\pm$0.51                          & 80.63$\pm$0.45 & 82.98$\pm$0.23       & 83.48$\pm$0.25                    & \textbf{84.59 $\pm$0.55} \\
    & 5             & 76.55$\pm$0.79                           & 80.44$\pm$0.74                           & 77.42$\pm$0.39 & 79.13$\pm$0.59                          & 78.39$\pm$0.54 & \textbf{82.27$\pm$0.45}         & 81.37$\pm$0.51                   & 81.22 $\pm$1.09  \\
    & 10            & 70.39$\pm$1.28                           & 75.61$\pm$0.59                           & 72.22$\pm$0.38 & 75.16$\pm$0.76                          & 71.47$\pm$0.83 & 79.03$\pm$0.59  & \textbf{83.68$\pm$0.23}  & 79.53 $\pm$1.60  \\
    & 15            & 65.10$\pm$0.71                           & 69.78$\pm$1.28                           & 66.82$\pm$0.39 & 71.03$\pm$0.64                          & 66.69$\pm$1.18 & 76.40$\pm$1.27  & 75.53$\pm$0.83  &\textbf{80.47 $\pm$0.71} \\
    & 20            & 59.56$\pm$2.72                           & 59.94$\pm$0.92                           & 59.27$\pm$0.37 & 65.71$\pm$0.89                          & 58.94$\pm$1.13 & 73.32$\pm$1.56  & 74.23$\pm$0.65  & \textbf{78.40 $\pm$2.18}   \\
    & 25            & 47.53$\pm$1.96                           & 54.78$\pm$0.74                           & 50.51$\pm$0.78 & 60.82$\pm$1.08                          & 52.06$\pm$1.19 & 69.72$\pm$1.69  & 64.74$\pm$0.83  & \textbf{78.99 $\pm$0.96} \\ \midrule
    \multirow{7}{*}{Citeseer} & 0             & 71.96$\pm$0.55                           & 73.26$\pm$0.83                           & 71.20$\pm$0.83 & 72.10$\pm$0.63                          & 70.65$\pm$0.32 & 73.28$\pm$0.69 & \textbf{76.29$\pm$0.63}  &  73.49 $\pm$1.96  \\
    & 5             & 70.88$\pm$0.62                           & 72.89$\pm$0.83                           & 70.50$\pm$0.43 & 70.51$\pm$0.97                          & 68.84$\pm$0.72 & 72.93$\pm$0.57            & \textbf{74.13$\pm$0.92}                  & 73.64 $\pm$1.33                    \\
    & 10            & 67.55$\pm$0.89                           & 70.63$\pm$0.48                           & 67.71$\pm$0.30 & 69.54$\pm$0.56                          & 68.87$\pm$0.62 & 72.51$\pm$0.75  & 71.89$\pm$1.04   & \textbf{72.73 $\pm$0.69}                    \\
    & 15            & 64.52$\pm$1.11                           & 69.02$\pm$1.09                           & 65.69$\pm$0.37 & 65.95$\pm$0.94                          & 63.26$\pm$0.96 & 72.03$\pm$1.11  & 72.09$\pm$1.32    & \textbf{73.02 $\pm$1.16}                      \\
    & 20            & 62.03$\pm$3.49                           & 61.04$\pm$1.52                           & 62.49$\pm$1.22 & 59.30$\pm$1.40                          & 58.55$\pm$1.09 & 70.02$\pm$2.28  & 66.09$\pm$1.05   &  \textbf{72.43 $\pm$1.48}                    \\
    & 25            & 56.94$\pm$2.09                           & 61.85$\pm$1.12                           & 55.35$\pm$0.66 & 59.89$\pm$1.47                          & 57.18$\pm$1.87 & 68.95$\pm$2.78  & 67.88$\pm$0.98    &  \textbf{73.19 $\pm$0.49}                   \\ \midrule
    \multirow{7}{*}{Polblogs} & 0             & 95.69$\pm$0.38 & 95.35$\pm$0.20                           & 95.22$\pm$0.14 & -                                       & 95.31$\pm$0.18 & -                             & -               & \textbf{95.67 $\pm$0.36}                    \\
    & 5             & 73.07$\pm$0.80                           & 83.69$\pm$1.45                           & 74.34$\pm$0.19 & -                                       & \textbf{89.09$\pm$0.22} & -                               & -             & 79.18 $\pm$1.12                   \\
    & 10            & 70.72$\pm$1.13                           & 76.32$\pm$0.85                           & 71.04$\pm$0.34 & -                                          & \textbf{81.24$\pm$0.49} & -                          & -               & 74.95 $\pm$1.08                      \\
    & 15            & 64.96$\pm$1.91                           & 68.80$\pm$1.14                           & 67.28$\pm$0.38 & -                                       & 68.10$\pm$3.73 & -                 & -                          & \textbf{70.14 $\pm$1.45}                      \\
    & 20            & 51.27$\pm$1.23                           & 51.50$\pm$1.63                           & 59.89$\pm$0.34 & -                                       & 57.33$\pm$3.15 & -                       & -                     & \textbf{65.85 $\pm$1.38}                    \\
    & 25            & 49.23$\pm$1.36                           & 51.19$\pm$1.49                           & 56.02$\pm$0.56 & -                                       & 48.66$\pm$9.93 & -                     & -                     & \textbf{63.37 $\pm$2.03}                  
    \\ \bottomrule
\end{tabular}}
\label{table1}
\end{table}

\subsubsection{Under the targeted adversarial attack}
\begin{figure}[htbp]
\centering
\subfigure[]{
\begin{minipage}[t]{1\textwidth}
  \centering
  \includegraphics[width = 0.6\linewidth]{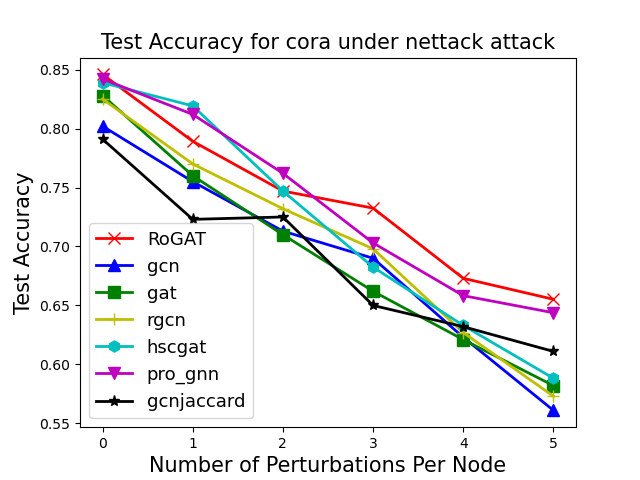}
  
\end{minipage}\label{cora}}
\subfigure[]{
\begin{minipage}[t]{1\textwidth}
  \centering
  \includegraphics[width = 0.6\linewidth]{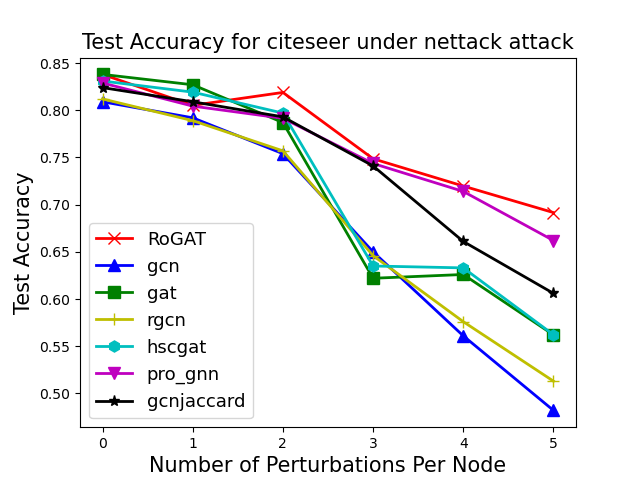}
  
\end{minipage}\label{citeseer}}
\subfigure[]{
\begin{minipage}[t]{1\textwidth}
\centering
\includegraphics[width = 0.6\linewidth]{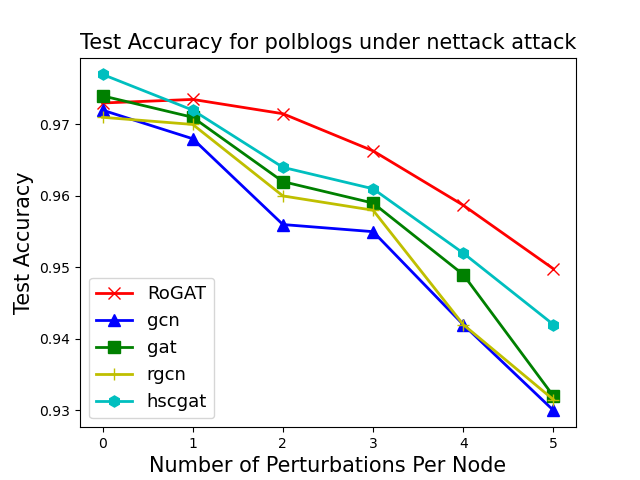}

\end{minipage}
\label{polblogs}}
\caption{Results of Cora, Citeseer, Polblogs under netttack}
\end{figure}

In this part, we evaluate the performance of different methods for node classification problems against the targeted attacks, which aim to attack selected nodes. Here we choose the nettack as the targeted-attack method and use the default parameter in the original paper~\cite{zugner2018adversarial}. The number of  perturbations per node varies from 1 to 5. And similar to~\cite{jin2020graph}, all the nodes with a degree larger than 10 are chosen as the targeted nodes. We display the performance of node classification for different methods. In Figure \ref{cora} and \ref{citeseer}, it shows that our method outperforms  most  methods and has a similar performance with Pro-GNN for the Cora and Citeseer. Our approach has 10\% and 20\% improvement, respectively, in Cora and Citeseer compared with the original GCN. Since the dataset polblogs  do not have node features, our method performs better than other methods except for GCN-SVD. 

%

\begin{figure}[htbp]
\centering
\subfigure[]{
\begin{minipage}[t]{1\textwidth}
  \centering
  \includegraphics[width = 0.6\linewidth]{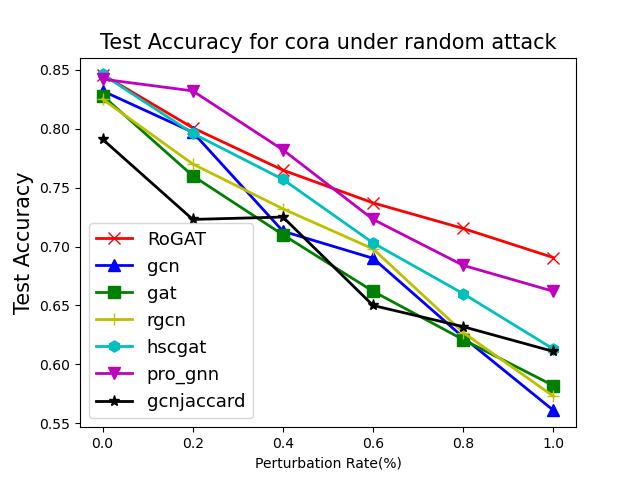}
  \label{cora2}
\end{minipage}}
\subfigure[]{
\begin{minipage}[t]{1\textwidth}
  \centering
  \includegraphics[width = 0.6\linewidth]{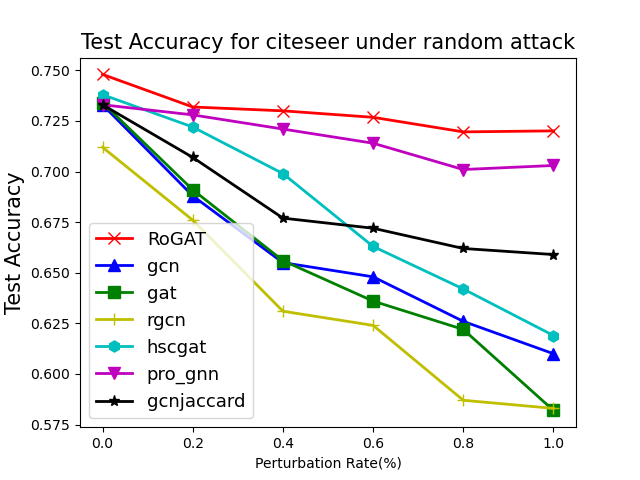}
\label{citeseer2}
\end{minipage}}
\subfigure[]{
\begin{minipage}[t]{1\textwidth}
  \centering
  \includegraphics[width = 0.6\linewidth]{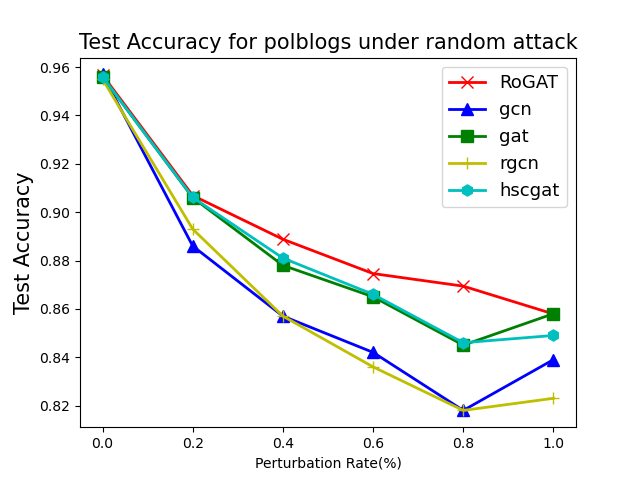}
\label{polblogs2}
\end{minipage}}
\caption{Results of Cora, Citeseer, Polblogs under random attack}
\end{figure}

We also do some experiments to evaluate our methods when dealing with the random attack, which adds the perturbation on nodes randomly. Different ratios of perturbations varying from 0\% to 100\% are adopted to disturb the graph structure. The result in Figure \ref{cora2} and \ref{citeseer2} shows that our RoGAT outperforms other methods in dealing with Cora and Citeseer and has relatively better performance with the dataset Polblogs. RoGAT has more than 15\% and 13\% improvement with Cora and Citeseer. It means that RoGAT can successfully resist the random attack.
The results in Figure \ref{cora2} and Figure \ref{citeseer2} show that our RoGAT performs significantly better than other methods when dealing with Cora and Citeseer datasets, with an accuracy  improvement of 15\% and 13\%, respectively. Since Polblogs do not have exact node features, RoGAT has relatively better performance, only slightly inferior to the gcn-svd method  in Figure \ref{polblogs2}.
Overall, RoGAT has quite good performance compared with most defensive methods when dealing with different types of adversarial attacks.

\subsection{Ablation study}
In this part, we build ablation studies to figure out how different components affect the model. As is shown in our model, two main pre-process procedures occur before training RoGAT. To understand the different impacts of each procedure, we eliminate the modification of graph structure and feature respectively to check the performance changes. Here we only report results on Cora, since other datasets have a similar performance.
We use RoGAT(no structure) and RoGAT(no feature) to represent RoGAT without modification structure and feature respectively.
In Table \ref{table_abdation}, we observe that when the attack is in low-level, RoGAT with no modification of structure has good performance, while RoGAT with no modification of feature has relatively better performance when the graph is heavily poisoned.
\begin{table}[htbp]
\caption{Classfication accuracy performance of RoGAT variants under metattack}
\begin{equation*}
\begin{array}{c|cccccc}\hline \text{ptb} & 0 & 0.05 & 0.10 & 0.15 &0.20 &0.25\\ \hline \text { RoGAT(no structure) } & 82.95 & 78.92 & 76.86 & 68.31 & 63.22&60.40\\ \text { RoGAT(no feature) } & 77.41 & 76.11 & 75.40 & 75.35 & 74.95&74.60

\\   \text { RoGAT} & \textbf{84.59} & \textbf{81.22} & \textbf{79.53} & \textbf{80.47} &\textbf{78.40}& \textbf{78.90}\\ \hline\end{array}
\end{equation*}
\label{table_abdation}
\end{table}

\subsubsection{Hyperparameter analysis}
\begin{figure*}
\centering
\begin{minipage}[b]{1\linewidth}
  \centering
  \includegraphics[width=0.6\linewidth]{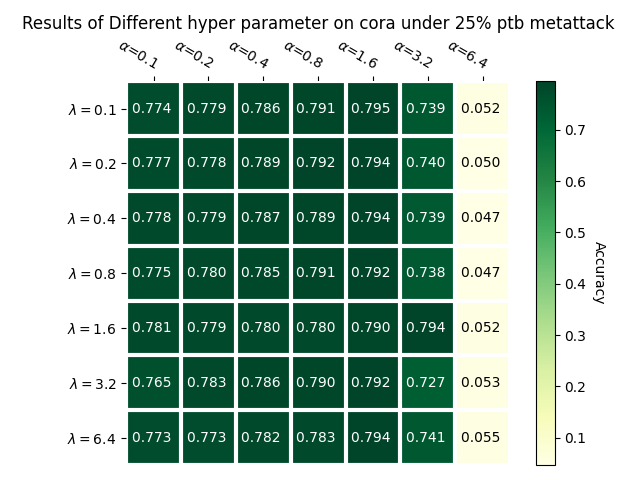}
  \caption*{Cora}
\end{minipage}
\begin{minipage}[b]{1\linewidth}
  \centering
\includegraphics[width=0.6\linewidth]{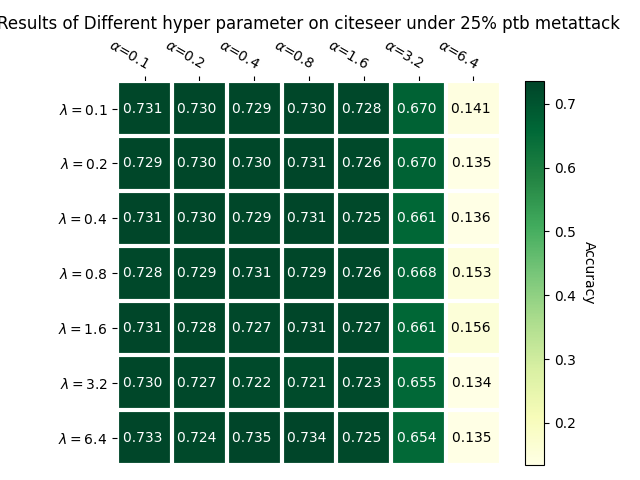}
    \caption*{Citeseer}
\end{minipage}
\begin{minipage}[b]{1\linewidth}
  \centering
\includegraphics[width=0.6\linewidth]{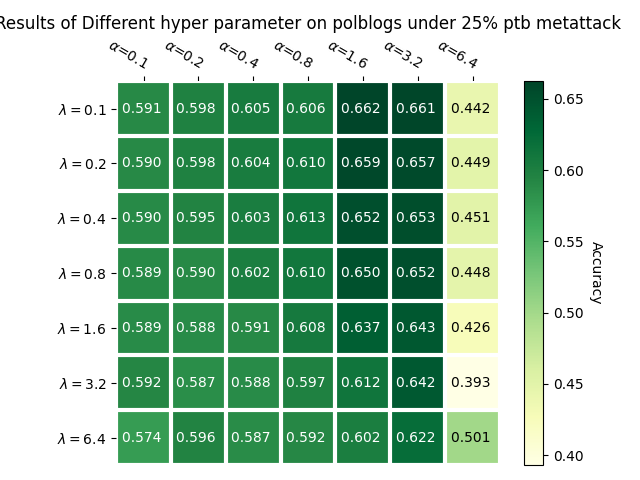}
    \caption*{Polblogs}
\end{minipage}
\caption{The performance for different $\lambda$ and $\gamma$ under 25\% ptb metattack(Green color represents the performance)}	\label{heatmap}
\end{figure*}

In this section, we discuss the influence of hyper-parameters  for  RoGAT.  Here we set $\alpha=\gamma$ and only consider the impact of $\alpha$ and $\lambda$ on the Cora dataset with perturbation rates of 25\% metattack. We vary  $\alpha$ and $\lambda $ from 0.1 to 6.4 in a log scale base 2 on the Cora dataset, respectively. 
Figure \ref{heatmap} shows the  test accuracy of RoGAT with different $\lambda$ and $\alpha$.  The introduction of $\lambda$ and $\alpha$ can contribute to the robustness of GAT.  And compared with $\lambda$,  the appropriate value of $\alpha$ has more influence on the performance of RoGAT. The performance of RoGAT is not sensitive to $\alpha$ and $\lambda$ with not too large $\alpha$. It means that the feature smoothness is tightly connected with the performance of RoGAT. For different kinds of datasets, $\alpha$ decides the ratio between two parts loss, which thus needs to be selected carefully. Therefore for RoGAT, using feature smoothness to revise the structure's attention is effective in defending adversarial attacks.

\subsection{Attention weight analysis}
Since we assume that the revised weight $\bar{\alpha}$ gives more attention to positive edges and reduces the influence of negative edges, we compute the ratio of weight between the negative edges and positive edges during the training procedure. Here we set $\alpha=\gamma=1$ and compute the average weight $\bar{\alpha}$ during the training procedure under the metattack. The fake edges represent the edges generated by adversarial attacks, while the real edges represent the edges in the original graph. 
For different rates of perturbations and datasets, the weight ratio between fake edges and real edges decreases from the initial value $1$ to a smaller value. 
And almost for all three datasets, the ratio decreases faster for the smaller perturbations, which leads to better performance. RoGAT can adjust the  ratio of contributions for fake and real edges. Therefore, it is consistent with the assumption that less attention will be given to the fake edges to reduce the influence of negative information during the aggregation procedure.
\begin{figure*}
\label{ratio}
\centering
\subfigure[]{
\begin{minipage}[b]{1\linewidth}
  \centering
  \includegraphics[width=0.6\linewidth]{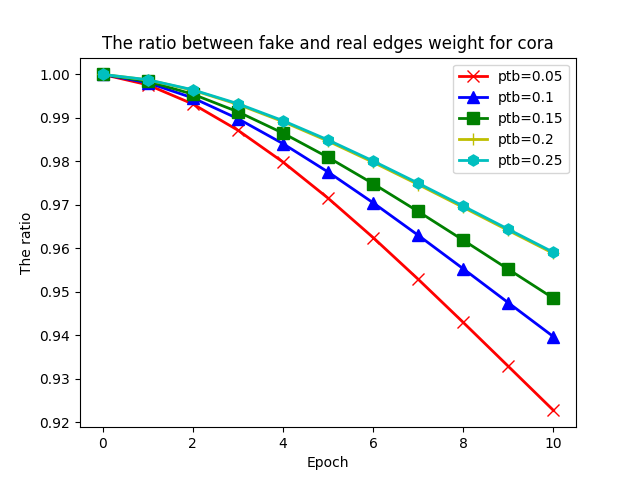}
\end{minipage}
}
\subfigure[]{
\begin{minipage}[b]{1\linewidth}
  \centering
  \includegraphics[width=0.6\linewidth]{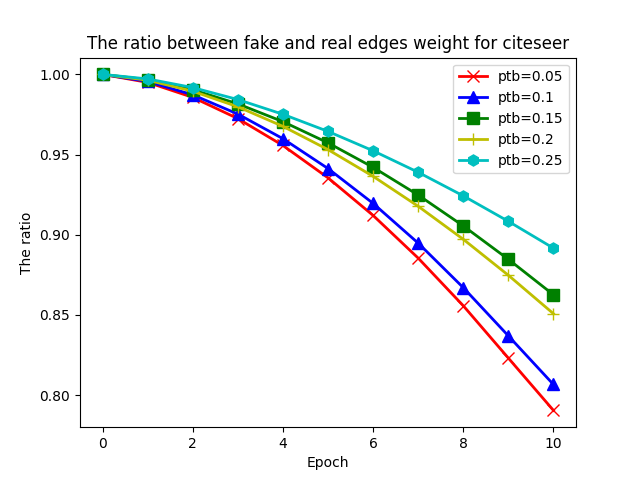}
\end{minipage}}
\subfigure[]{
\begin{minipage}[b]{1\linewidth}
  \centering
  \includegraphics[width=0.6\linewidth]{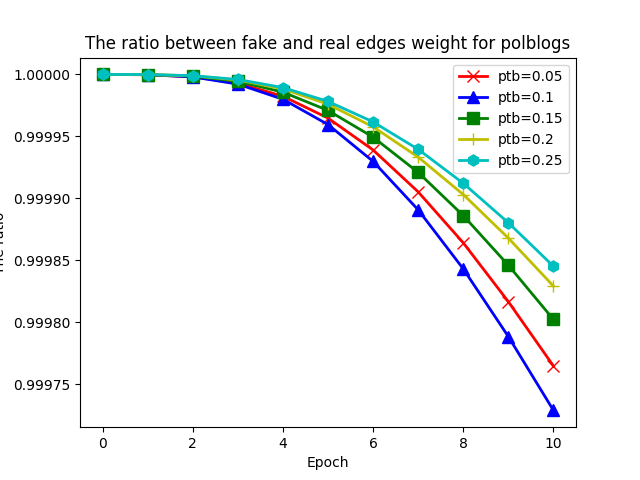}
\end{minipage}}
\caption{The ratio of the weight of fake and real edges during training procedure for cora, citeseer and Polblogs}
\end{figure*}

\section{Conclusion}
Graph neural networks, including graph convolutional networks and graph attention networks, are easily disturbed by graph adversarial attacks. This paper adjusts the attention mechanism and then proposes the robust GAT called RoGAT, which revises the structure and feature of the poisoned graph iteratively. The results of experiments show that RoGAT can reduce the influence of fake edges and performs better than most of the recent baselines, especially in defending the metattack.  Therefore, the prior information can help us to revise the attention score for fake and real edges progressively. Different graphs including homogeneous and heterogeneous have different prior information, which deserves further research in designing robust algorithms.

\section*{Acknowlegements}
This research work is supported by the National Science Foundation of China(NSFC) under 61977065.

\bibliography{ref.bib}
\end{document}